\documentclass[11pt]{article}

\usepackage[margin=1in]{geometry}
\usepackage[T1]{fontenc}
\usepackage[utf8]{inputenc}
\usepackage{lmodern}
\usepackage{amsmath, amssymb}
\usepackage{booktabs}
\usepackage{enumitem}
\usepackage{graphicx}
\usepackage{float}
\usepackage{multirow}
\usepackage{hyperref}
\usepackage{makecell}

\title{Automated Instruction Revision (AIR):\\
A Structured Comparison of Task Adaptation Strategies for LLM Systems}
\author{%
  \textbf{Solomiia Bilyk$^{1}$} \qquad
  \textbf{Volodymyr Getmanskyi$^{1}$} \qquad
  \textbf{Taras Firman$^{1, 2}$} \\
  \\
  $^{1}$ Eleks Ltd.\\
  \texttt{\{solomiia.bilyk, volodymyr.getmanskyi, taras.firman\}@eleks.com} \\
  \\
  $^{2}$ Ukrainian Catholic University, Lviv, Ukraine \\ \texttt{firman@ucu.edu.ua}
}
\date{\today}

\begin{document}

\maketitle

\begin{abstract}
This paper studies Automated Instruction Revision (AIR), a rule-induction-based method for adapting large language models (LLMs) to downstream tasks using limited task-specific examples. We position AIR within the broader landscape of adaptation strategies, including prompt optimization, retrieval-based methods, and fine-tuning. We then compare these approaches across a diverse benchmark suite designed to stress different task requirements, such as knowledge injection, structured extraction, label remapping, and logical reasoning. The paper argues that adaptation performance is strongly task-dependent: no single method dominates across all settings. Across five benchmarks, AIR was strongest or near-best on label-remapping classification, while KNN retrieval performed best on closed-book QA, and fine-tuning dominated structured extraction and event-order reasoning. AIR is most promising when task behavior can be captured by compact, interpretable instruction rules, while retrieval and fine-tuning remain stronger in tasks dominated by source-specific knowledge or dataset-specific annotation regularities.
\end{abstract}


\section{Introduction}

Large language models (LLMs) are being adopted across a growing range of research and industry workflows, but adapting them reliably to a concrete downstream task remains difficult. In practice, the challenge is not only selecting a base model, but also choosing how that model should be adapted to the task at hand. A practitioner may rely on manual prompt iteration, retrieval over demonstrations or external sources, automated prompt optimization, or parameter-updating approaches such as fine-tuning. Each option offers a different tradeoff in data requirements, latency, interpretability, engineering cost, and robustness.

This adaptation decision is often still mediated by repeated expert effort. Prompt engineers and data scientists must inspect failures, rewrite instructions, curate examples, and tune task-specific heuristics before a system behaves acceptably. Such expert-in-the-loop iteration can be effective, but it is expensive, hard to standardize, and difficult to scale across many tasks. One motivation for this paper is therefore to reduce dependence on this repeated manual intervention by making task adaptation more structured and more data-driven.

At the same time, existing adaptation approaches do not admit a universal winner. Retrieval-based methods are often attractive when performance depends on access to source-specific knowledge or highly relevant exemplars at inference time. Fine-tuning can absorb dataset-specific regularities and decision boundaries into model parameters, but typically requires more infrastructure and offers less transparency about what changed. Prompt-based strategies are lightweight and flexible, yet their success often depends on brittle manual iteration or opaque search procedures. These tradeoffs suggest that the best adaptation strategy depends strongly on the structure of the task rather than on a single globally superior method.

We study this question through Automated Instruction Revision (AIR), a rule-induction and refinement pipeline for deriving task-specific guidance from limited examples. Instead of storing adaptation knowledge only in retrieved exemplars or in updated weights, AIR aims to represent useful task behavior as explicit instruction rules that can be iteratively proposed, filtered, and refined. This makes AIR appealing as an interpretable alternative when compact instructions can summarize task behavior.

To evaluate this claim, we compare AIR against representative alternatives spanning prompt-based adaptation, retrieval-based methods, and fine-tuning across a diverse benchmark suite. The benchmarks are chosen to stress qualitatively different task settings, including label remapping, structured extraction, knowledge-sensitive prediction, and reasoning over constrained patterns. Our goal is not to identify a single universally best method, but to characterize when AIR is competitive and when other approaches are a better fit.

Our results suggest that the choice of adaptation strategy should be task-dependent. AIR is most effective when task behavior can be compressed into interpretable instruction rules, while retrieval and fine-tuning remain stronger when success depends primarily on stored knowledge, source-specific evidence, or annotation regularities that are difficult to summarize explicitly.

\section{Related Work}

Work on improving large language model performance can be grouped into several closely related directions. One line of research addresses the quality of data used for adaptation. In particular, prior studies have examined automatic revision of instruction-tuning data and the incorporation of few-shot-style guidance directly into training, especially in low-resource settings~\cite{wang2023coachlm,fewshotfinetune2025}. The common idea behind these efforts is that model behavior can be improved not only through inference-time prompt design, but also through stronger supervision signals and more effective adaptation strategies during training.

A second direction concerns automatic prompt engineering. Early work demonstrated that prompts can be generated and selected automatically rather than crafted only by humans~\cite{zhou2022ape}. Subsequent methods extended this idea in different ways: PRewrite formulates prompt rewriting as an optimization problem~\cite{singh2024prewrite}, while other approaches investigate how different prompts may be better suited to different groups of inputs~\cite{zhang2024autopromptselection}. Taken together, these studies frame prompt construction as a systematic process of search and optimization rather than a purely manual activity.

Another important direction moves beyond optimizing a single prompt and instead considers broader LLM systems. Within this line of work, DSPy models LLM applications as multi-stage programs composed of interacting modules~\cite{khattab2023dspy}, while MiPRO targets the optimization of both instructions and few-shot demonstrations for such programs~\cite{opsahlong2024mipro}. More recent contributions extend this system-level perspective further. TextGrad relies on iterative textual feedback to improve prompts and other text-based components~\cite{yuksekgonul2024textgrad}, GEPA refines prompts through reflective updates based on system behavior~\cite{gepa2025}, and Maestro broadens the optimization scope by modifying not only prompts and parameters but also the agent graph itself~\cite{maestro2025}. Collectively, these approaches reflect a shift from isolated prompt tuning toward optimization of full multi-component LLM systems.

Alongside prompt and system optimization, another related stream of research explores interpretable structure induction with LLMs. This tendency is especially visible in work on decision trees. Existing studies investigate whether LLMs can generate explicit tree-like decision structures from task descriptions and feature information, and how such structures can connect LLM reasoning with decision-tree-based modeling~\cite{zan2025zeroshottree,dimitrov2024pleasegivemeatree,llmmeeting2025}. These works are relevant to our setting because they pursue a similar objective: rather than relying solely on implicit model behavior, they aim to recover explicit, human-readable decision logic.

Overall, the literature reveals two main tendencies. The first is the movement from manual prompt writing toward automated optimization of prompts, examples, and full LLM pipelines. The second is the growing interest in interpretable decision structures such as rules or trees. Our work lies at the intersection of these directions: like prompt optimization methods, it seeks automatic improvement of LLM behavior, but like tree-based approaches, it emphasizes the discovery of explicit and reusable decision logic.

\section{AIR: Automated Instruction Revision}

\subsection{Method Overview}
We propose Automated Instruction Revision (AIR), a data-driven prompt adaptation pipeline that learns explicit task guidance from labeled examples instead of relying on continuous manual prompt engineering. The central idea is to transform supervised task data into a compact set of task-specific decision rules, convert these rules into an executable system prompt, and iteratively refine them using additional sampled cases from the training set.

\subsection{Pipeline}
The AIR pipeline proceeds through clustering, rule induction, rule aggregation, and targeted refinement. Starting from labeled training data, it constructs an executable instruction set in the following stages.

\begin{enumerate}[leftmargin=1.8em]
  \item \textbf{Standardize the task representation and compute embeddings.} AIR first maps each dataset into canonical input and output columns and computes embeddings for both sides of the supervision signal. This creates a uniform representation across tasks and provides the structure used later for clustering and local comparison.

\item \textbf{Form semantically local but output-diverse clusters.} AIR first clusters the input embeddings with KMeans. The number of clusters \(K\) is treated as a hyperparameter and is set to the default value of 5 in our experiments. This gives an initial semantic grouping of the inputs, together with the distance from each sample to every cluster centroid. The outputs are processed separately: if the output representations behave like discrete labels, AIR maps them directly to categories; otherwise, it clusters the output embeddings with KMeans using the same \(K\). AIR then adjusts the initial input-cluster assignments to make the clusters more informative for output discrimination. For each sample, it checks nearby input clusters in order of distance and prefers a reassignment only if adding that sample makes the cluster’s output distribution closer to, or at least not farther from, the overall output distribution in the dataset. This closeness is measured by the mean squared error between the local cluster distribution and the global one. If no suitable alternative is found, the sample keeps its original nearest-cluster assignment. Finally, AIR performs a repair step that redistributes samples from clusters containing only one output class to nearby alternative clusters. This encourages groups to remain semantically coherent while still preserving output variation that is useful for distinguishing behavior.

  \item \textbf{Induce local contrastive rules.} Within each final cluster, AIR constructs balanced A/B example sets drawn from different output groups. These localized contrasts are passed to a reasoning model with an instruction to infer a small number of rules of the form \emph{if condition on the input, then output action or pattern}. The purpose is not free-form explanation, but extraction of compact decision-boundary rules that distinguish competing output behaviors within the same semantic neighborhood.

\item \textbf{Aggregate and compile the rules into an executable prompt.} Because rule induction is repeated across clusters, the initial rule pool is usually large and partially redundant. AIR therefore applies a dedicated LLM-based rule-compilation stage. In this stage, a reasoning model receives the induced rules and, following a constrained compiler prompt, groups rules with semantically similar \texttt{THEN} actions, identifies the shared structure of their \texttt{IF} conditions, merges complementary signals, removes lexical noise, and preserves only the exclusions needed to avoid cross-rule collisions. The aggregated rules produced by this step are then concatenated with the task description to form a structured final prompt that instructs the model to follow the decision process encoded in the rules. AIR also creates a traced variant of this prompt that asks the model to return both the final prediction and the identifiers of the applied rules, enabling later refinement.

  \item \textbf{Refine rules iteratively on fresh sampled cases.} Rather than using the aggregated rules as final, AIR re-evaluates them on newly sampled examples and records predicted outputs, applied rule identifiers, and task metric values. For each rule, the pipeline separates participating cases into mistakes, where the prediction was incorrect, and anchors, where the prediction was correct according to the provided metric. Small refinement batches built from these two sets are then sent back to the reasoning model, which is asked to make the smallest necessary local revision to the current rule. In this way, AIR updates individual rules while attempting to preserve behavior on successful anchor cases.
\end{enumerate}

\begin{figure}[H]
  \centering
  \includegraphics[width=0.95\textwidth]{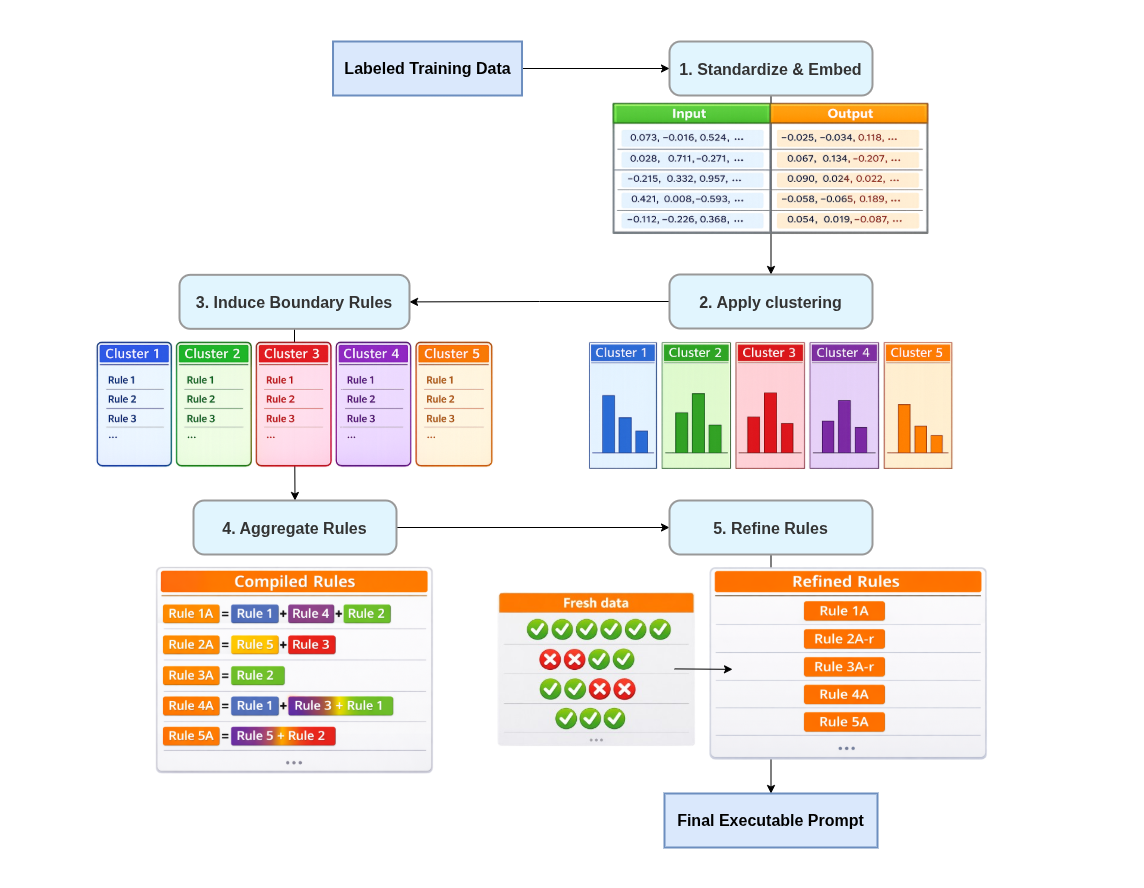}
  \caption{Overview of the AIR pipeline from labeled task data to compiled and refined instructions.}
  \label{fig:air-pipeline}
\end{figure}

After refinement, the updated rules are assembled into the final system prompt used for downstream evaluation. In this way, AIR derives task-specific instructions from labeled examples through rule induction, aggregation, and revision.  Figure~\ref{fig:air-pipeline} summarizes the full pipeline and highlights how these stages are connected.

\subsection{Strengths and Limitations}
AIR is intended to provide several practical advantages. First, it is interpretable: the learned adaptation is represented as readable instruction text rather than hidden inside parameter updates. Second, it can support rule-level inspection and revision: individual rules can be examined and modified without changing model weights. Third, the pipeline preserves intermediate artifacts that can help analyze how the final prompt was constructed. Fourth, it reduces manual effort by automating a process that would otherwise require a person to inspect the data, identify recurring patterns, and summarize them as explicit rules.

At the same time, AIR has important limitations. Its core assumption is that useful task behavior can be externalized as natural-language guidance. When labels are inconsistent, examples are noisy, or correct decisions depend on latent patterns that are difficult to verbalize, rule induction may become unstable. AIR can also suffer from rule interaction effects: a locally beneficial instruction may conflict with others after aggregation, and repeated revisions may reduce clarity instead of improving it.

For these reasons, we treat AIR not as a universal solution to task adaptation, but as a structured and interpretable option for settings in which task behavior can be captured, at least in part, by explicit instruction rules derived from supervision.

\section{Experimental Setup}
This section describes the benchmark suite, the compared adaptation methods, and the models used in our experiments. The evaluation is designed to support a controlled comparison of prompting, retrieval-based adaptation, prompt optimization, and fine-tuning under a shared task-specific setup.

\subsection{Benchmarks}
We evaluate AIR on a deliberately diverse benchmark suite chosen to reflect different sources of adaptation difficulty rather than a single task family. The suite spans remapped-label classification, closed-book factual QA, schema-constrained extraction, PII identification, and event-order reasoning. Across all benchmarks, data are partitioned into train, development, and test splits. The training split is used as the main adaptation set, including fine-tuning, AIR fitting, few-shot construction, and KNN retrieval support. The development split is used for validation-time optimization and prompt selection for methods such as GEPA, MIPROv2, and TextGrad. The test split remains held out until the final evaluation.

For \textbf{classification}, we use real customer-support requests collected from Twitter based on the Customer Support dataset~\cite{customer_support_twitter}, from which we construct an 8-class benchmark by selecting requests addressed to eight companies.  To suppress memorized brand priors, explicit company mentions and other direct lexical cues are removed from the request text. We also reassign the output labels so that each company is consistently mapped to a different description rather than its original one. The resulting task forces the model to infer the latent label mapping from content alone rather than from world knowledge or familiar brand associations. The final dataset is balanced across the selected companies. Performance is measured with exact match. The dataset was split into 168/8/64 train/dev/test examples, respectively.

For \textbf{closed-book question answering}, we build a benchmark from \emph{Ever Young} by Alice Gerstenberg~\cite{gerstenberg_ever_young} and create question-answer pairs that must be answered without providing the source text at inference time. The benchmark is intentionally designed so that the relevant source material is unlikely to be part of the model's pretrained knowledge. We initially considered combining multiple plays, but ultimately focused on one richer source to keep the benchmark more controlled and internally consistent. To evaluate outputs, we use an LLM-as-a-judge rubric with four dimensions: correctness, completeness, absence of hallucination, and focus, each scored on a 0/0.5/1 scale, with the final score defined as the average of the four subscores. The dataset was split into 100/5/30 train/dev/test examples, respectively.

For \textbf{information extraction}, we use campaign-finance filings for Philadelphia elections from the City of Philadelphia Campaign Finance Reports source~\cite{phila_campaign_finance} containing transaction-level records for contributions, expenditures, and debt. Each example is transformed from a structured table row into a raw CSV-style string consisting of a header and one data line, and the columns are randomly shuffled. This means the model must first reconstruct the field mapping before it can extract the requested outputs. In addition to recovering directly stated fields, the model must infer two derived labels: \texttt{candidate\_relationship}, which categorizes how the donor relates to the candidate, and \texttt{contribution\_context}, which combines donor type and contribution size into labels. Performance is measured as average exact match across the two target fields. The dataset was split into 210/12/78 train/dev/test examples, respectively.

For \textbf{PII extraction}, we use PUPA (Private User Prompt Annotations), a benchmark derived from real WildChat user-assistant conversations that contain explicit PII leakage~\cite{papillon_2025}. In the original setting, PUPA is part of a broader privacy-aware delegation pipeline that balances answer quality against privacy exposure. In our benchmark, however, we isolate only the annotation component and convert it into a pure extraction task: given \texttt{user\_query}, the model must output the gold \texttt{pii\_units} directly, without rewriting, redaction, or response generation. Evaluation is performed with entity-level exact-match F1. The dataset was split into 180/60/60 train/dev/test examples, respectively.

For \textbf{event logical reasoning}, we adopt the Event Logic Reasoning subset of BizFinBench.v2~\cite{bizfinbench_v2}. Each example describes a finance-related scenario involving multiple market events, and the task is to recover their correct logical order, either temporally or through cause-effect structure. The benchmark is intended to test whether this ordering logic can be induced through prompting, transferred from similar retrieved examples, or absorbed through fine-tuning. The output is a compact event-index sequence, for example ``2,1,4,3,'' rather than a free-form explanation. Performance is measured with exact match. The dataset was split into 180/60/60 train/dev/test examples, respectively.

\subsection{Compared Methods}
We compare AIR against a set of adaptation strategies implemented under a shared task-specific workflow across our benchmarks. All methods start from the same manually written task instruction and use the available labeled training data, but they differ in whether they adapt the prompt, retrieve examples, or update model parameters.

\begin{itemize}[leftmargin=1.5em]
  \item \textbf{Initial prompt baseline.} The model is evaluated with the manually written task prompt only, without retrieval, optimization, or parameter updates. This baseline measures the performance of direct zero-shot prompting with human-authored instructions.

  \item \textbf{KNN-based prompting.} For each test example, we retrieve similar labeled training instances and append them as in-context examples to the initial prompt. This baseline tests whether instance-level adaptation through dynamic example selection is sufficient without modifying the instruction itself.

  \item \textbf{DSPy BootstrapFewShot.} DSPy automatically compiles a few-shot prompt by sampling labeled training examples and retaining demonstrations that satisfy the task metric. This yields an automatically selected demonstration set while keeping the base instruction fixed.

  \item \textbf{Fine-tuning.} We fine-tune the base chat model on the task training split using chat-formatted supervision constructed from the same initial system prompt, the task input, and the gold output. The resulting model is then evaluated on held-out data. This baseline represents parameter adaptation rather than prompt-only adaptation.

  \item \textbf{DSPy MIPROv2.} MIPROv2 is used as a prompt optimization baseline without a separate teacher or reflection model. In our experiments, it is run with DSPy's \texttt{auto="medium"} budget and task-specific limits on bootstrapped and labeled demonstrations, which are chosen separately for each benchmark. It jointly searches over instruction candidates and, when enabled, demonstration candidates to find a high-performing compiled prompt.

  \item \textbf{DSPy GEPA.} GEPA is used as a reflective prompt optimizer with the same \texttt{auto="medium"} budget, but unlike MIPROv2 it relies on a separate reasoning model as a teacher for reflection. In our setup, GEPA receives task-specific textual feedback through the metric and uses the teacher model to revise candidate instructions based on execution traces and validation-time search.

  \item \textbf{TextGrad.} TextGrad treats the system prompt as an optimizable text variable while keeping the model parameters fixed. In our experiments, we run one epoch over the full training set; within each minibatch, per-example losses are summed into a single minibatch loss, which is then used to update the prompt. This baseline therefore represents iterative text-space optimization of the initial instruction.

  \item \textbf{AIR.} AIR induces explicit task-specific rules from labeled examples, aggregates them into a compact rule set, and refines them iteratively before assembling the final prompt.

\end{itemize}

\subsection{Models}
Across the benchmark suite, we use \texttt{gpt-4.1-mini-2025-04-14} as the base task model for evaluation. This model is used for the initial prompt baseline, KNN-based prompting, DSPy BootstrapFewShot, MIPROv2, and the final prompt evaluation in AIR, and it also serves as the starting point for fine-tuning. For methods that require a stronger auxiliary model for reflection or rule revision, we use \texttt{gpt-5-2025-08-07} as the teacher model in GEPA, TextGrad, and AIR. In AIR, we additionally use \texttt{text-embedding-3-small} to compute input and output embeddings for clustering and rule induction. For the closed-book question answering benchmark, the LLM-as-a-judge metric uses \texttt{gpt-5.1} as the judge model.

\section{Results}

This section reports the quantitative results for all compared methods across the five benchmark tasks. In addition to the task metric, each table includes the token usage reported in the original experiment slides, separating training-time token usage by model role and inference-time token usage for the base model.

\paragraph{Classification.}
Table~\ref{tab:results_classification} shows the classification results. GEPA achieves the highest score, while AIR remains very close and also surpasses fine-tuning. This pattern suggests that, once brand priors are removed, the task behaves less like ordinary text classification and more like learning a remapped latent label system, where explicit prompt-level instruction discovery can be highly effective.

\renewcommand{\arraystretch}{1.15}
\begin{table}[H]
\centering
\small
\caption{Classification (8 classes). Metric: Accuracy.}
\label{tab:results_classification}
\resizebox{\textwidth}{!}{%
\begin{tabular}{lccccc}
\toprule
\multirow{2}{*}{Method} &
\multirow{2}{*}{\makecell[c]{Metric\\(higher is better)}} &
\multicolumn{3}{c}{Train Tokens (input / output)} &
\multirow{2}{*}{\makecell[c]{Inference Tokens\\(input / output)\\Base Model}} \\
\cmidrule(lr){3-5}
& & Embed Model & Base Model & Reflection Model & \\
\midrule
Initial prompt               & 00.00 & --       & --               & --               & 6752 / 180 \\
KNN                          & 09.37 & --       & --               & --               & 9008 / 183 \\
Fine-tuning                 & 90.63 & --       & 54153 / 1362     & --               & 6752 / 179 \\
DSPy BootstrapFewShot (\#16) & 18.75 & --       & 48874 / 562      & --               & 80032 / 934 \\
DSPy MIPROv2                 & 31.25 & --       & 1676737 / 23626  & --               & 75257 / 886 \\
DSPy GEPA                    & \textbf{96.88} & --       & 1636241 / 10261  & 107506 / 94565   & 184569 / 879 \\
TextGrad                     & 46.87 & --       & 501625 / 394     & 876468 / 348127  & 475808 / 215 \\
AIR                          & \underline{95.31} & 7282 / - & 54704 / 1392     & 10301 / 10981    & 55392 / 174 \\
\bottomrule
\end{tabular}%
}
\end{table}

\paragraph{Closed-book question answering.}
The closed-book QA results in Table~\ref{tab:results_qa} show a different problem settings. KNN clearly performs best, while most prompt-optimization methods fail to improve over the initial prompt. This suggests that the benchmark is dominated by source-specific knowledge injection rather than by generic reasoning or procedural instruction improvement.

\renewcommand{\arraystretch}{1.15}
\begin{table}[H]
\centering
\small
\caption{Closed-book Q\&A. Metric: LLM.}
\label{tab:results_qa}
\resizebox{\textwidth}{!}{%
\begin{tabular}{lccccc}
\toprule
\multirow{2}{*}{Method} &
\multirow{2}{*}{\makecell[c]{Metric\\(higher is better)}} &
\multicolumn{3}{c}{Train Tokens (input / output)} &
\multirow{2}{*}{\makecell[c]{Inference Tokens\\(input / output)\\Base Model}} \\
\cmidrule(lr){3-5}
& & Embed Model & Base Model & Reflection Model & \\
\midrule
Initial prompt               & 42.08 & --       & --               & --               & 2123 / 608 \\
KNN                          & \textbf{81.67} & --       & --               & --               & 3321 / 434 \\
Fine-tuning                 & \underline{72.08} & --       & 36250 / 5945     & --               & 2123 / 319 \\
DSPy BootstrapFewShot (\#10) & 38.30 & --       & 7275 / 223       & --               & 21983 / 592 \\
DSPy MIPROv2                 & 37.50 & --       & 906850 / 74221   & --               & 14573 / 604 \\
DSPy GEPA                    & 50.83 & --       & 1098892 / 15248  & 118233 / 88819   & 58193 / 623 \\
TextGrad                     & 37.50 & --       & 405807 / 2925    & 601023 / 684425  & 127373 / 356 \\
AIR                          & 42.08 & 3878 / - & 123152 / 2281    & 19665 / 31183    & 68063 / 382 \\
\bottomrule
\end{tabular}%
}
\end{table}

\paragraph{Information extraction.}
Table~\ref{tab:results_info_extraction} shows that fine-tuning is dominant on the information-extraction task. This is consistent with the structure of the benchmark: before performing extraction, the model must first reconstruct the field mapping from shuffled CSV-style rows. AIR performs substantially worse here, which is consistent with the difficulty of capturing this task through compact local rules alone.

\renewcommand{\arraystretch}{1.15}
\begin{table}[H]
\centering
\small
\caption{Information extraction. Metric: Mean per field accuracy.}
\label{tab:results_info_extraction}
\resizebox{\textwidth}{!}{%
\begin{tabular}{lccccc}
\toprule
\multirow{2}{*}{Method} &
\multirow{2}{*}{\makecell[c]{Metric\\(higher is better)}} &
\multicolumn{3}{c}{Train Tokens (input / output)} &
\multirow{2}{*}{\makecell[c]{Inference Tokens\\(input / output)\\Base Model}} \\
\cmidrule(lr){3-5}
& & Embed Model & Base Model & Reflection Model & \\
\midrule
Initial prompt               & 36.54 & --         & --                & --                & 62099 / 1651 \\
KNN                          & \underline{55.12} & --         & --                & --                & 219450 / 1371 \\
Fine-tuning                 & \textbf{98.71} & --         & 502968 / 11178    & --                & 62099 / 1377 \\
DSPy BootstrapFewShot (\#12) & 42.95 & --         & 55479 / 144       & --                & 736331 / 1829 \\
DSPy MIPROv2                 & 42.31 & --         & 6399076 / 40216   & --                & 358187 / 1838 \\
DSPy GEPA                    & 53.21 & --         & 2142493 / 20824   & 206378 / 91251    & 265211 / 1847 \\
TextGrad                     & 25.64 & --         & 284301 / 1723     & 218746 / 194031   & 187991 / 1404 \\
AIR                          & 35.90 & 200111 / - & 176632 / 3559     & 102743 / 52487    & 172235 / 1658 \\
\bottomrule
\end{tabular}%
}
\end{table}

\paragraph{PUPA (PII).}
The PUPA results in Table~\ref{tab:results_pupa} again favor fine-tuning, with MIPROv2, AIR, and GEPA forming a second tier. This pattern suggests that the benchmark rewards not only general PII detection, but also adherence to dataset-specific annotation habits. AIR remains competitive with GEPA, but does not match the best parameter-adaptation result.

\renewcommand{\arraystretch}{1.15}
\begin{table}[H]
\centering
\small
\caption{PUPA (PII). Metric: F1.}
\label{tab:results_pupa}
\resizebox{\textwidth}{!}{%
\begin{tabular}{lccccc}
\toprule
\multirow{2}{*}{Method} &
\multirow{2}{*}{\makecell[c]{Metric\\(higher is better)}} &
\multicolumn{3}{c}{Train Tokens (input / output)} &
\multirow{2}{*}{\makecell[c]{Inference Tokens\\(input / output)\\Base Model}} \\
\cmidrule(lr){3-5}
& & Embed Model & Base Model & Reflection Model & \\
\midrule
Initial prompt               & 17.58 & --        & --                & --               & 19263 / 2683 \\
KNN                          & 35.08 & --        & --                & --               & 34052 / 2965 \\
Fine-tuning                 & \textbf{68.48} & --        & 190881 / 7086     & --               & 19263 / 611 \\
DSPy BootstrapFewShot (\#3)  & 44.23 & --        & 16281 / 237       & --               & 37889 / 1475 \\
DSPy MIPROv2                 & \underline{64.91} & --        & 1490233 / 134750  & --               & 52409 / 1480 \\
DSPy GEPA                    & 59.29 & --        & 1321431 / 229379  & 24659 / 13374    & 71669 / 1574 \\
TextGrad                     & 55.54 & --        & 1298401 / 14733   & 161048 / 177955  & 123063 / 986 \\
AIR                          & 59.32 & 90088 / - & 329236 / 2451     & 55382 / 57279    & 252543 / 621 \\
\bottomrule
\end{tabular}%
}
\end{table}

\noindent\emph{Note.} In the source results, AIR on PUPA was terminated after 8 steps, including 5 consecutive steps with no improvement, using batch size 3.

\paragraph{BizFinBench.v2 (Event Logical Reasoning).}
Finally, Table~\ref{tab:results_bizfinbench} shows that fine-tuning again performs best on event logical reasoning. The relatively strong initial-prompt score suggests that the base model already has some latent capability for financial event ordering, while fine-tuning appears to stabilize how that reasoning is mapped into the required output sequence. KNN and AIR both provide moderate gains, but neither approaches the fine-tuned model.

\renewcommand{\arraystretch}{1.15}
\begin{table}[H]
\centering
\small
\caption{BizFinBench.v2 (Event Logical Reasoning). Metric: Accuracy.}
\label{tab:results_bizfinbench}
\resizebox{\textwidth}{!}{%
\begin{tabular}{lccccc}
\toprule
\multirow{2}{*}{Method} &
\multirow{2}{*}{\makecell[c]{Metric\\(higher is better)}} &
\multicolumn{3}{c}{Train Tokens (input / output)} &
\multirow{2}{*}{\makecell[c]{Inference Tokens\\(input / output)\\Base Model}} \\
\cmidrule(lr){3-5}
& & Embed Model & Base Model & Reflection Model & \\
\midrule
Initial prompt               & 45.00 & --        & --                & --               & 28477 / 1036 \\
KNN                          & \underline{51.67} & --        & --                & --               & 41249 / 895 \\
Fine-tuning                 & \textbf{73.34} & --        & 247908 / 7560     & --               & 28477 / 856 \\
DSPy BootstrapFewShot (\#3)  & 35.00 & --        & 17001 / 344       & --               & 72338 / 1456 \\
DSPy MIPROv2                 & 43.33 & --        & 1616492 / 38339   & --               & 72278 / 1456 \\
DSPy GEPA                    & 46.67 & --        & 1998619 / 23064   & 37111 / 26706    & 170138 / 1396 \\
TextGrad                     & 43.33 & --        & 728850 / 78210    & 7589 / 112655    & 51157 / 976 \\
AIR                          & \underline{51.67} & 65923 / - & 453192 / 3008     & 58716 / 65583    & 292237 / 1036 \\
\bottomrule
\end{tabular}%
}
\end{table}

\section{Discussion}

The results do not support a single universal winner across all benchmarks. Instead, they show that different methods are strongest under different task conditions. Retrieval works best when the task is dominated by source-specific knowledge, fine-tuning is strongest when the task depends on stable dataset-specific mappings or output conventions, and instruction-based adaptation is most competitive when the target behavior can be expressed as explicit decision rules.

Within this picture, AIR occupies a clear practical niche. It does not win on every benchmark, but it often improves substantially over the initial prompt and remains especially competitive on tasks where the target behavior can be captured as reusable rules. This makes AIR attractive in settings where interpretability is important and some loss relative to the best task-specific method is acceptable.

AIR also appears to offer a better efficiency--interpretability trade-off than more expensive prompt-optimization methods such as GEPA. Although GEPA achieves the best result on classification, AIR remains close while requiring fewer calls to the reasoning model. This suggests that AIR can provide much of the benefit of instruction discovery at lower optimization cost.

Another important point is that the compared adaptation methods still rely on a separate search or optimization stage to identify a strong task-specific solution. When the data changes, this process may need to be repeated rather than updated incrementally. In principle, AIR may offer an advantage here, since new data could potentially be incorporated by extending or refining the rule set instead of rebuilding the adaptation from scratch. However, this possibility was not evaluated in the present experiments and remains a direction for future work.

At the same time, the weaker AIR results help clarify its current limits. Its performance drops when the task depends more on factual injection, latent structure reconstruction, or dataset-specific annotation habits than on compact explicit rules. Overall, the results suggest that AIR is a promising middle-ground approach: it is not a replacement for retrieval or fine-tuning, but it offers a strong trade-off when explicit, interpretable task guidance is desired.

\section{Conclusion}

This paper presented Automated Instruction Revision (AIR), a prompt adaptation pipeline that derives explicit task-specific rules from labeled examples, compiles them into an executable instruction set, and refines them iteratively through additional evaluation cases. AIR was evaluated against direct prompting, retrieval-based prompting, fine-tuning, and several prompt-optimization baselines across five diverse benchmarks.

The main empirical result is not that AIR dominates the comparison, but that different adaptation methods are favored by different problem settings. Retrieval performs best when success depends on source-specific factual information, fine-tuning is strongest when the task requires stable structural behavior or dataset-specific conventions, and instruction-focused methods are most effective when the target behavior can be captured through explicit decision rules. AIR is therefore best understood as a structured alternative for tasks where the required distinctions must be learned from the provided task data rather than inferred from prior model knowledge. It is less effective when performance depends more on source-specific facts or unstable input structure than on reusable explicit rules.

Overall, the study supports a more focused claim: effective LLM adaptation is strongly task-dependent, and AIR should be viewed as a promising but still developing approach to explicit instruction learning from data. Compared with optimization-heavy baselines such as GEPA or TextGrad, AIR is also comparatively less demanding in terms of computational budget, as it requires fewer repeated calls to a stronger reasoning model during the adaptation process. At the same time, the results suggest that its current limitations arise less from rule induction itself than from the difficulty of maintaining rule quality during aggregation and refinement, making these stages the most important directions for future work.

\section{Future Work}

The results indicate that AIR is promising, especially on tasks where adaptation can be externalized into compact and interpretable rules, but they also make clear that several parts of the pipeline would benefit from further development. A first priority is improving rule merging, as the aggregation stage remains vulnerable to partial overlap, contradictions, paraphrased redundancy, and local noise, limiting the consistency and clarity of the final prompt.

The role of clustering also remains unresolved. AIR currently uses clustering to construct semantically local comparison sets before rule induction, but the present study does not establish when this step is necessary, most beneficial, or safely omitted. This question matters because clustering introduces both computational cost and additional design choices. 

The refinement stage likewise requires closer study. If sampled batches do not reflect the full dataset distribution, refinement may overfit local patterns rather than improve overall task behavior. More representative sampling, or strategies that adapt dynamically to coverage gaps, could make AIR's iterative updates more reliable.

More broadly, several pipeline hyperparameters are still selected heuristically. Important settings such as the number of clusters and the number of refinement iterations should be chosen more systematically. Evaluation should also be extended to additional LLM families in order to clarify how robust AIR is across different base-model capabilities, instruction-following behavior, and reasoning styles.

\bibliographystyle{plain}

\end{document}